%% file: 0_main.tex
\documentclass{article}

\usepackage{microtype}
\usepackage{graphicx}
\usepackage{booktabs} 

\usepackage{hyperref}

\usepackage{wrapfig}

\usepackage[accepted]{icml2024}

\usepackage{amsmath}
\usepackage{amssymb}
\usepackage{mathtools}
\usepackage{amsthm}

\usepackage{balance}

\usepackage[capitalize,noabbrev]{cleveref}

\theoremstyle{plain}

\theoremstyle{definition}

\theoremstyle{remark}

\usepackage[textsize=tiny]{todonotes}
\usepackage{caption}

\usepackage{microtype}
\RequirePackage{amsmath,amsfonts,amssymb,bm,latexsym}
\usepackage{subfig}
\usepackage{float}
\usepackage{multirow}
\usepackage{tikz}
\usepackage{enumitem}
\usetikzlibrary{fadings}
\usetikzlibrary{mindmap,trees}
\usetikzlibrary{arrows,automata,shapes,positioning,shadows,trees}
\usetikzlibrary{decorations.pathmorphing}
\usetikzlibrary{shapes,shadows}
\usetikzlibrary{shapes.arrows}
\usetikzlibrary{calc,shapes, positioning}
\usetikzlibrary{decorations.text}
\usetikzlibrary{matrix,chains,positioning,decorations.pathreplacing,arrows}
\usetikzlibrary{bayesnet}
\usepackage[edges]{forest}
\usetikzlibrary{shadows.blur}
\usetikzlibrary{shapes.geometric}

\usepackage{pgfplots}
\usepackage{pgfplotstable}
\usepackage{pgfmath,pgffor}
\usepgfplotslibrary{colorbrewer}
\pgfplotsset{compat = 1.14, cycle list/Set1-8}
\usetikzlibrary{pgfplots.statistics, pgfplots.colorbrewer}
\pgfplotsset{compat=1.8}

\tikzstyle{edge}=[-latex',draw=black!90,shorten <=1pt,shorten >=1pt]
\tikzstyle{redge}=[latex'-,draw=black!90,shorten <=1pt,shorten >=1pt]
\tikzstyle{dedge}=[latex'-latex',draw=black!90,shorten <=1pt,shorten >=1pt]

\tikzstyle{block}=[draw, text width=5em,align=center,shape=rectangle, rounded corners, , align=center]
\tikzstyle{nobox}=[align=center]
\definecolor{emb}{RGB}{209,228,252}
\definecolor{hidden-blue}{RGB}{194,232,247}
\definecolor{hidden-orange}{RGB}{224,224,224}
\definecolor{hidden-yellow}{RGB}{242,244,193}
\definecolor{output-purple}{RGB}{219,203,231}
\definecolor{output-green}{RGB}{204,231,207}
\definecolor{output-blue}{RGB}{44,169,225}

\definecolor{output-black}{RGB}{0,0,0}
\definecolor{output-white}{RGB}{255,255,255}
\definecolor{myblue}{RGB}{137,195,235}

\definecolor{hiddendraw}{RGB}{137,195,235}

\tikzstyle{emb-purple}=[
    rectangle,
    draw=output-purple!50!purple,
    fill=output-purple,
    text opacity=1,
    minimum height=1.5em,
    minimum width=1.5em,
    inner sep=0pt,
    align=center,
    fill opacity=.5,
    ]

\tikzstyle{emb-blue}=[
    rectangle,
    draw=emb!50!blue,
    fill=emb,
    text opacity=1,
    minimum height=1.5em,
    minimum width=1.5em,
    inner sep=0pt,
    align=center,
    fill opacity=.5,
]

\makeatletter
\renewcommand\paragraph{\@startsection{paragraph}{4}{\z@}%
{1ex \@plus1ex \@minus.2ex}%
 {-1em}%
 {\normalfont\normalsize\bfseries}}
\makeatother

\usetikzlibrary{arrows.meta,shapes,positioning,shadows,trees}
\newcommand{\secref}[1]{Section~\ref{#1}}

\usepackage{todonotes}
\usepackage{xcolor}

\usepackage{tcolorbox}
\definecolor{rq}{HTML}{1B365C}
\definecolor{rqBack}{HTML}{9ECBF7}

\icmltitlerunning{What Can Large Language Models Tell Us about Time Series Analysis}

\begin{document}

\twocolumn[

\icmltitle{Position: What Can Large Language Models Tell Us about Time Series Analysis}



\icmlsetsymbol{equal}{*}
\icmlsetsymbol{correspondence}{\dag}

\begin{icmlauthorlist}
\icmlauthor{Ming Jin}{aaa,equal}
\icmlauthor{Yifan Zhang}{bbb,equal}
\icmlauthor{Wei Chen}{ccc,equal}
\icmlauthor{Kexin Zhang}{ddd}
\icmlauthor{Yuxuan Liang}{correspondence,ccc}
\icmlauthor{Bin Yang}{eee} \\
\icmlauthor{Jindong Wang}{fff}
\icmlauthor{Shirui Pan}{correspondence,aaa}
\icmlauthor{Qingsong Wen}{correspondence,hhh}
\end{icmlauthorlist}

\icmlaffiliation{aaa}{Griffith University.}
\icmlaffiliation{bbb}{Chinese Academy of Sciences.}
\icmlaffiliation{ccc}{The Hong Kong University of Science and Technology (Guangzhou).}
\icmlaffiliation{ddd}{Zhejiang University.}
\icmlaffiliation{eee}{East China Normal University.}
\icmlaffiliation{fff}{Microsoft Research Asia.}
\icmlaffiliation{hhh}{Squirrel AI}

\icmlcorrespondingauthor{Yuxuan Liang}{yuxliang@outlook.com}
\icmlcorrespondingauthor{Shirui Pan}{s.pan@griffith.edu.au}
\icmlcorrespondingauthor{Qingsong Wen}{qingsongedu@gmail.com}

\icmlkeywords{Machine Learning, ICML}

\vskip 0.3in 

]



\printAffiliationsAndNotice{\icmlEqualContribution} 

\input{1_abstract}

\input{2_introduction}

\input{3_background}

\input{4_methodology}

\input{5_outlook_conclusion}




\clearpage
\input{7_statements}
\balance
\bibliography{8_reference}
\bibliographystyle{icml2024}

\clearpage
\balance
\appendix
\onecolumn

\input{6_appendix}

\end{document}

%% file: 1_abstract.tex
\begin{abstract}
Time series analysis is essential for comprehending the complexities inherent in various real-world systems and applications. Although large language models (LLMs) have recently made significant strides, the development of artificial general intelligence (AGI) equipped with time series analysis capabilities remains in its nascent phase. Most existing time series models heavily rely on domain knowledge and extensive model tuning, predominantly focusing on prediction tasks. In this paper, we argue that current LLMs have the potential to revolutionize time series analysis, thereby promoting efficient decision-making and advancing towards a more universal form of time series analytical intelligence. Such advancement could unlock a wide range of possibilities, including  time series modality switching and question answering. We encourage researchers and practitioners to recognize the potential of LLMs in advancing time series analysis and emphasize the need for trust in these related efforts. Furthermore, we detail the seamless integration of time series analysis with existing LLM technologies and outline promising avenues for future research.
\end{abstract}

%% file: 2_introduction.tex
\vspace{-0.3cm}
\section{Introduction}

\begin{figure}
    \centering
    \includegraphics[width=1 \linewidth]{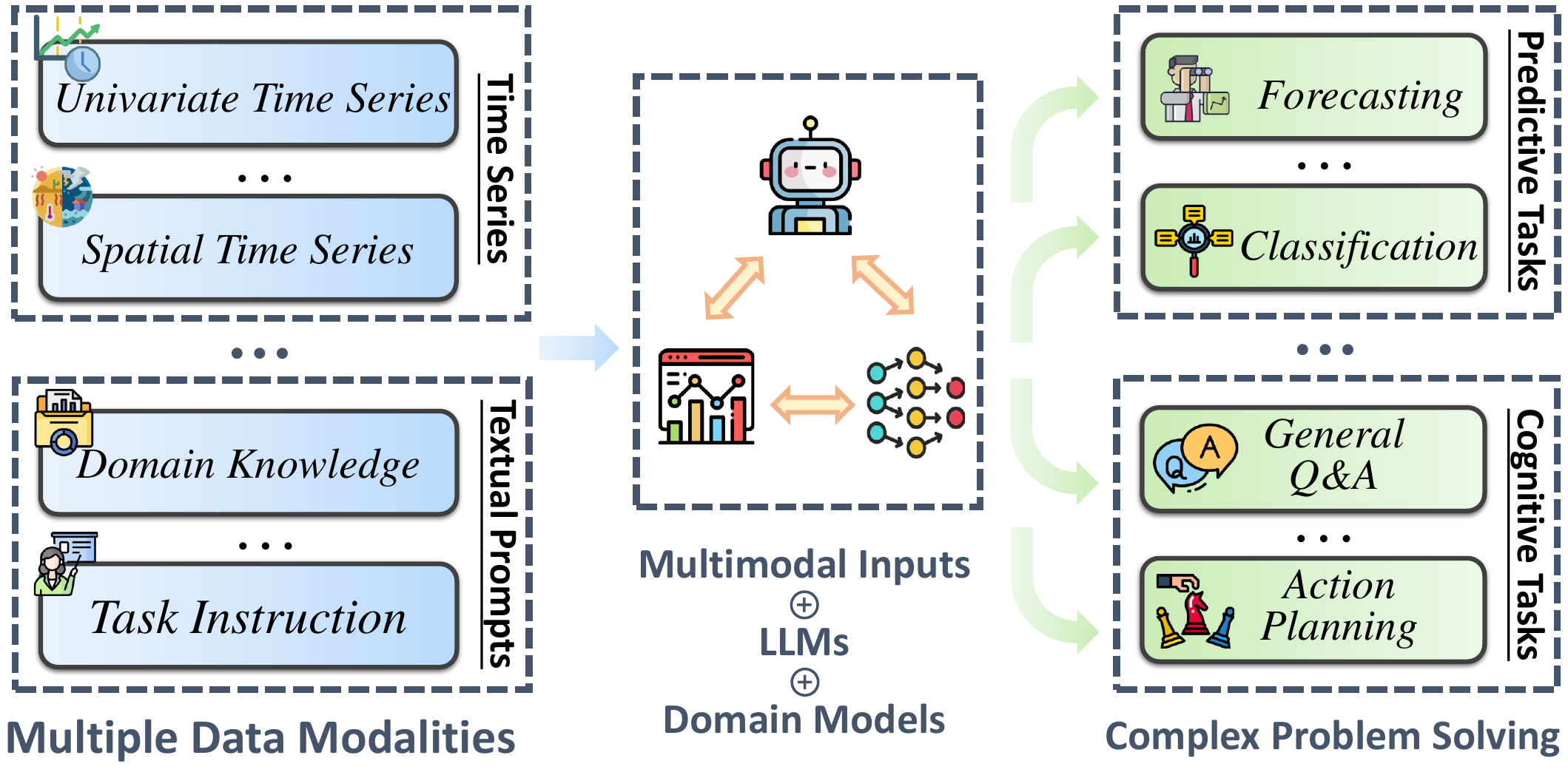}
    \vspace{-5mm}
    \caption{Across a myriad of time series analytical domains, the integration of time series and LLMs demonstrates potential in solving complex real-world problems.}
    \label{fig:llm4ts}
    \vspace{-5mm}
\end{figure}

Time series, a fundamental data type for recording dynamic system variable changes, is widely applied across diverse disciplines and applications~\cite{hamilton2020time,wen2022robust}. Its analysis is instrumental in uncovering patterns and relationships over time, thus facilitating the understanding of complex real-world systems and supporting informed decision-making. Many real-world dynamic laws, such as financial market fluctuations~\cite{tsay2005analysis} and traffic patterns during peak hours~\cite{alghamdi2019forecasting}, are fundamentally encapsulated in time series data. In addressing practical scenarios, time series analysis employs methods ranging from traditional statistics~\cite{fuller2009introduction} to recent deep learning techniques~\cite{gamboa2017deep,wen2020time}. In the era of sensory artificial intelligence, these domain-specific models efficiently extract meaningful representations for prediction tasks like forecasting and classification. Despite such successes, a notable gap persists between mainstream time series research and the development of artificial general intelligence (AGI)~\cite{bubeck2023sparks} with time series capabilities to address various problems in a unified manner.

The recent emergence of large language models (LLMs), such as Llama~\cite{touvron2023llama, touvron2023llama2} and GPT-4~\cite{achiam2023gpt}, have swept through and propelled advancements in various interdisciplinary fields~\cite{zhao2023survey}. Their outstanding zero-shot capabilities~\cite{kojima2022large}, along with emerging reasoning and planning abilities~\cite{wang2023survey}, have garnered increasing attention. However, their focus has primarily been on text sequences. The exploration of extending LLMs' capabilities to accommodate and process more data modalities, such as images~\cite{zhang2023vision} and graphs~\cite{chen2023exploring}, has begun to receive preliminary attention.

With the integration of LLMs, time series analysis is undergoing significant transformation~\cite{jin2023large,liang2024foundation}. Time series models are conventionally designed for specific tasks, depend heavily on prior domain knowledge and extensive model tuning, lacking assurances of effective updates and validations~\cite{zhou2023one1}. Conversely, LLMs hold enormous potential not only to improve prediction performance~\cite{jin2023time} but also to support cross-disciplinary~\cite{yan2023urban}, interactive~\cite{xue2023weaverbird}, and interpretative~\cite{gu2023anomalygpt} analyses. By aligning time series and natural language, large language and foundation/specialistic time series models constitute a new technology paradigm, where the LLM is prompted with both time series and textual instructions. In this paradigm, time series and textual information provide essential contexts, LLMs contribute internal knowledge and reasoning capabilities, and pre-trained time series models offer fundamental pattern recognition assurances. This novel integration is depicted in Figure \ref{fig:llm4ts}, where the successful amalgamation of these components showcases the potential for a general-purpose, unified system in next-generation time series analysis.

\begin{figure}
    \centering
    \includegraphics[width=0.5 \linewidth]{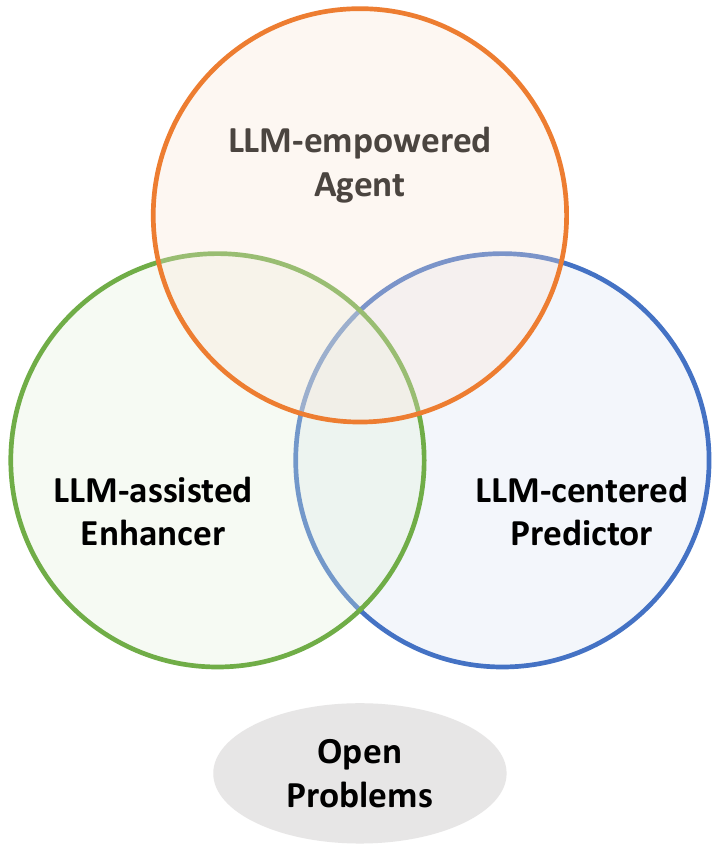}
    \vspace{-1mm}
    \caption{Task-solving capability boundaries on the roles of LLMs for time series analysis: as data/model enhancers, effective predictors, or next-generation agents.}
    \label{fig:relations}
    \vspace{-3mm}
\end{figure}

\paragraph{Why This Position Paper?} Given the remarkable capabilities emerging in recent research~\cite{jin2023large,liang2024foundation,zhang2024large}, we believe that the field of time series analysis research is undergoing an exciting transformative moment. Our standpoint is that LLMs can act as the central hub for understanding and advancing the analysis of time series data. Specifically, we present key insights that LLMs can profoundly impact time series analysis in three fundamental ways with their capability boundaries illustrated in Figure \ref{fig:relations}: \emph{\textbf{(1) as effective data and model enhancers}}, augmenting time series data and existing approaches with enhanced external knowledge and analytical prowess; \emph{\textbf{(2) as superior predictors}}, utilizing their extensive internal knowledge and emerging reasoning abilities to benefit a range of prediction tasks; and \emph{\textbf{(3) as next-generation agents}}, transcending conventional roles to actively engage in and transform time series analysis. We advocate attention to related research and efforts, moving towards more universal intelligent systems for general-purpose time series analysis. To this end, we thoroughly examine relevant literature, present and discuss potential formulations of LLM-centric time series analysis to bridge the gap between the two. We also identify and outline prospective research opportunities and challenges, calling for greater commitment and exploration in this promising interdisciplinary field.

\paragraph{Contributions:} The contributions of this work can be summarized in three aspects: 
\emph{\textbf{(1) offering new perspectives.}} We articulate our stance on LLM-centric time series analysis, outlining the potential synergies between LLMs and time series analytical models. This underscores the need for increased research focus and dedication in this area;
\emph{\textbf{(2) systematic review and categorization.}} We meticulously examine existing preliminary work and present a clear roadmap, highlighting three potential integration forms of LLMs and time series analysis;
\emph{\textbf{(3) identifying future opportunities.}} We explore and articulate areas that current research has not yet addressed, presenting promising directions for future investigations in this evolving interdisciplinary field.

%% file: 3_background.tex
\section{Background}

\subsection{Time Series Analysis}

\paragraph{Data Modality.} 
Time series data, comprising sequential observations over time, can be either regularly or irregularly sampled, with the latter often leading to missing values. This data falls into two main categories: \emph{univariate} and \emph{multivariate}. Univariate time series consist of single scalar observations over time, represented as $X = \{ x_1, x_2, \cdots, x_T \} \in \mathbb{R}^{T}$. Multivariate time series, on the other hand, involve $N$-dimensional vector observations, denoted as $X \in \mathbb{R}^{N \times T}$. In complex real-world systems, multivariate time series often exhibit intricate spatial dependencies in addition to temporal factors. This has led to some recent studies modeling them as graphs~\cite{jin2023survey}, also referred to as \emph{spatial time series}. In this approach, a time series is conceptualized as a sequence of graph snapshots, $\mathcal{G} = \{ \mathcal{G}_1, \mathcal{G}_2, \cdots, \mathcal{G}_T \}$, with each $G_t = (A_t, X_t)$ representing an attributed graph characterized by an adjacency matrix $A_t \in \mathbb{R}^{N \times N}$ and node features $X_t \in \mathbb{R}^{N \times D}$.

\begin{figure*}[htbp]
    \centering
    \includegraphics[width=0.8\linewidth]{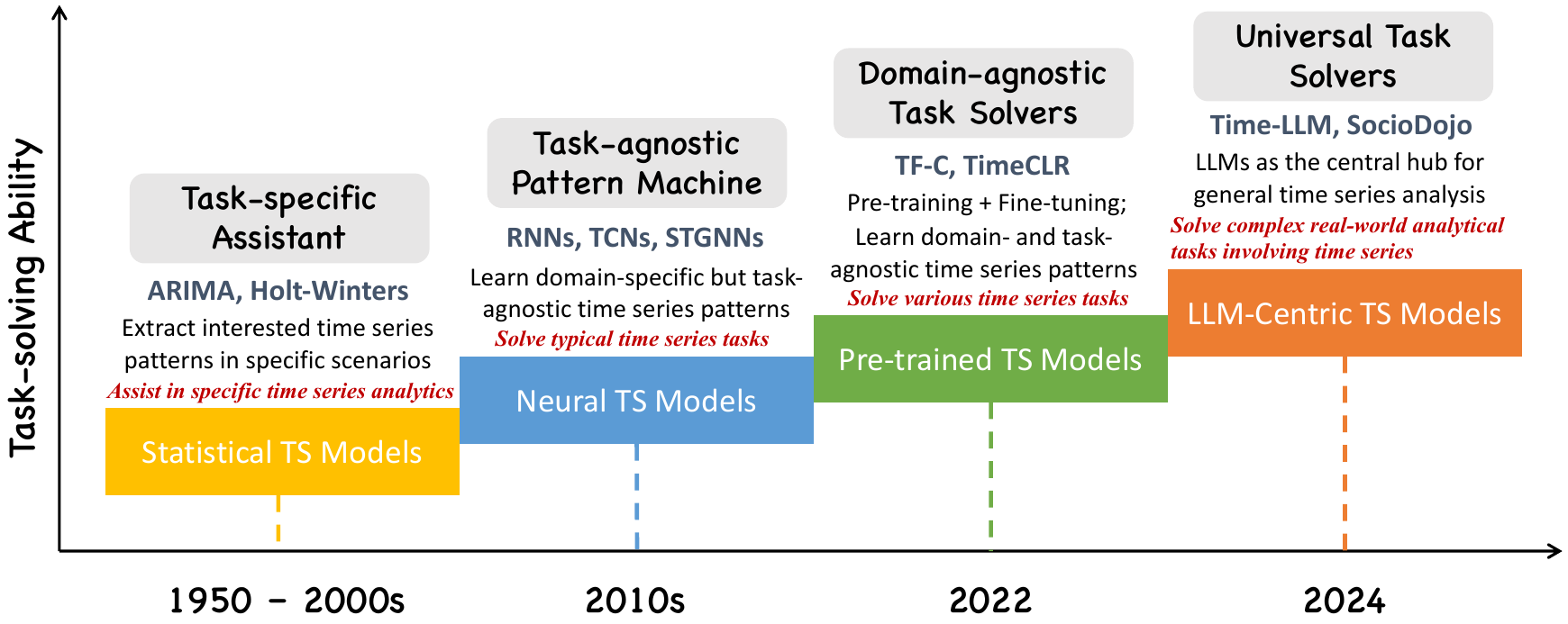}
    \vspace{-2mm}
    \caption{A roadmap of time series analysis delineating four generations of models based on their task-solving capabilities.}
    \label{fig:roadmap}
\end{figure*}

\paragraph{Analytical Tasks.}
Time series analysis is crucial for deriving insights from data, with recent deep learning advancements spurring a rise in neural network-based methods~\cite{wen2023transformers}.
These methods focus on modeling complex inter-temporal and/or inter-variable relationships in time series~\cite{zhang2023self,jin2023large}, aiding in tasks like forecasting, classification, anomaly detection, and imputation. Forecasting predicts future values, classification categorizes series by patterns, anomaly detection identifies anomalous events, and imputation estimates missing data.
Emerging research has also shown promise in time series modality switching and question answering~\cite{xue2023promptcast,jin2023time,yang2022zero}. These novel approaches highlight the potential for cross-disciplinary, interactive, and interpretative advancements in time series analytics. Such advancements open a realm of possibilities in practical applications, such as (zero-shot) medical question answering~\cite{yu2023zero,oh2023ecg} and intelligent traffic agents~\cite{da2023open,lai2023large}.

\subsection{Large Language Models}
\paragraph{Basic Concept.} 
Large language models typically refer to transformer-based pre-trained language models (PLMs) with billions or more parameters. The scaling of PLMs, both in terms of model and data size, has been found to enhance model performance across various downstream tasks~\cite{zhao2023survey}. These models such as GPT-4~\cite{achiam2023gpt}, PaLM~\cite{chowdhery2023palm}, and Llama~\cite{touvron2023llama}, undergo extensive pre-training on extensive text corpora, enabling them to acquire wide-ranging knowledge and problem-solving capabilities for diverse NLP tasks. Technically, language modeling (LM) is a fundamental pre-training task in LLMs and a key method for advancing machine language intelligence. The primary objective of LM is to model the probability of generating word sequences, encompassing both non-autoregressive and autoregressive language model categories. Autoregressive models, like the GPT series~\cite{bubeck2023sparks}, predict the next token $y$ based on a given context sequence $X$, trained by maximizing the probability of the token sequence given the context:
\begin{equation}
    P(y \mid X)=\prod_{t=1}^{T} P\left(y_{t} \mid x_{1}, x_{2}, \ldots, x_{t-1}\right),
\end{equation}
where $T$ represents the sequence length. Through this, the model achieves intelligent compression and language generation in an autoregressive manner.

\paragraph{Emergent Abilities of LLMs.}
Large language models exhibit emergent abilities that set them apart from traditional neural networks. These abilities, present in large models but not in smaller ones, are a significant aspect of LLMs~\cite{wei2022emergent}. Three key emergent abilities of LLMs include: \textit{(1) in-context learning (ICL)}, introduced by GPT-3~\cite{brown2020language}, allowing LLMs to generate relevant outputs for new instances using instructions and examples without additional training; \textit{(2) instruction following}, where LLMs, through instruction tuning, excel at novel tasks presented in an instructional format, enhancing their generalization~\cite{sanh2021multitask}; \textit{(3) step-by-step reasoning}, where LLMs use strategies like chain-of-thought (CoT)~\cite{wei2022chain} or other prompting strategies~\cite{yao2023tree, besta2023graph} to address complex tasks requiring multiple reasoning steps.

\subsection{Research Roadmap}
Time series analytical model development spans four generations: \emph{(1) statistical models}, \emph{(2) deep neural networks}, \emph{(3) pre-trained models}, and \emph{(4) LLM-centric models}, as shown in Figure~\ref{fig:roadmap}. This categorization hinges on the evolving task-solving capabilities of each model generation. 
Traditional analytics relied on statistical models like ARIMA~\cite{shumway2017arima} and Holt-Winters~\cite{kalekar2004time}, optimized for small-scale data and based on heuristics like stationarity and seasonality~\cite{hamilton2020time}. These models assumed past trends would continue into the future. Deep neural networks, like recurrent and temporal convolution neural networks~\cite{gamboa2017deep}, processed larger, complex datasets, capturing non-linear and long-term dependencies without heavy reliance on prior knowledge, thus transforming predictive time series analysis. Recent research like TimeCLR~\cite{yeh2023toward} introduced pre-training on diverse, large-scale time series data, allowing fine-tuning for specific tasks with relatively smaller data samples~\cite{jin2023large}, reducing the time and resources required for model training. This allows for the application of sophisticated models in scenarios where collecting large-scale time series data is challenging. Despite the successes of previous generations, we posit that the emergence of LLMs is set to revolutionize time series analysis, shifting it from predictive to general intelligence. LLM-centric models, processing both language instructions and time series~\cite{jin2023time, anonymous2024sociodojo}, extend capabilities to general question answering, interpretable predictions, and complex reasoning, moving beyond conventional predictive analytics.

%% file: 4_methodology.tex
\section{LLM-assisted Enhancer for Time Series}\label{sec:enhancer}
The role of LLM-assisted enhancers delves into whether LLMs can augment our understanding of time series data and extend the knowledge of existing models. Numerous methods have been devised to address temporal data, but the vast internal knowledge and reasoning capabilities of LLMs may significantly enhance both data understanding and model performance; thus, we intuitively distinguish LLM-assisted enhancers from \emph{data} and \emph{model} perspectives.

\subsection{Data-based Enhancer}\label{sec:input}
LLM-assisted enhancers not only enhance data interpretability but also provide supplementary improvements, facilitating a more thorough understanding and effective use of time series data.
For interpretability, LLMs offer textual descriptions and summaries, helping to understand patterns and anomalies in time series data. Examples include LLM-MPE~\cite{liang2023exploring} for human mobility data, SignalGPT~\cite{liu2023biosignal} for biological signals, and Insight Miner~\cite{zhang2023insight} for trend mining. Additionally, AmicroN~\cite{chatterjee2023amicron} and SST~\cite{ghosh2023spatio} use LLMs for detailed sensor and spatial time series analysis.
Supplementary enhancements involve integrating diverse data sources, enriching time series data context and improving model robustness, as explored in \cite{yu2023temporal} and \cite{fatouros2024can} for financial decision-making. Such enhancements help improve domain models' inherent capabilities and make them more robust.

\subsection{Model-based Enhancer}\label{sec:embedding}
Model-based enhancers aim to augment time series models by addressing their limitations in external knowledge and domain-specific contexts. Transferring knowledge from LLMs boosts the performance of domain models in handling complex tasks. Such approaches often employ a dual-tower model, like those in \cite{qiu2023automated, li2023frozen}, use frozen LLMs for electrocardiogram (ECG) analysis. Some methods further utilize contrastive learning to achieve certain alignments. For example, IMU2CLIP~\cite{moon2023imu2clip} aligns text and video with sensor data, while STLLM~\cite{anonymous2024st} enhances spatial time series prediction. 
Another branch utilizes prompting techniques to harness the inferential decision-making capability of LLMs. For instance, TrafficGPT~\cite{zhang2023trafficgpt} exemplifies decision analysis, integrating traffic models with LLMs for user-tailored solutions, offering detailed insights to enhance system interpretability.

\subsection{Discussion}
LLM-assisted enhancers effectively address the inherent sparsity and noise characteristics of time series data while also providing existing time series models with enhanced external knowledge and analytical capabilities. 
Time series, perhaps more than other modalities, benefits significantly from data enhancements due to its potentially lower information density. For example, comparing raw data volumes, video data inherently contain vastly more information per second than typical time series data such as audio. This disparity underscores the importance of time series data augmentation in many use cases, where LLMs can leverage both internal and external knowledge, along with their inherent reasoning capabilities, to address this challenge. On the other hand, the development of foundation models for time series is still in its infancy, presenting an opportunity for LLMs to extend the knowledge boundaries of existing time series models.
Moreover, this technology is plug-and-play, enabling flexible assistance for real-world time series data and model challenges. However, a notable hurdle is that using LLM as an enhancer introduces significant time and cost overheads when dealing with large-scale datasets. In addition, the inherent diversity and range of application scenarios in time series data add layers of complexity to the creation of universally effective LLM-assisted enhancers.

\begin{tcolorbox}[top=1pt, bottom=1pt, left=1pt, right=1pt]
  \textbf{Our position:}~\textit{LLM-assisted enhancers represent a promising avenue for augmenting time series data and models, meriting further exploration. Future directions should focus on developing efficient, accountable, and universally adaptable plug-and-play solutions that effectively address practical challenges, such as data sparsity and noise, while also considering the time and cost efficiencies for large-scale dataset applications.}
\end{tcolorbox} 

\section{LLM-centered Predictor for Time Series}\label{sec:predictor}

LLM-centered predictors utilize the extensive knowledge within LLMs for diverse time series tasks such as prediction and anomaly detection. Adapting LLMs to time series data involves unique challenges such as differences in data sampling and information completeness. In the following discussion, approaches are categorized into \emph{tuning-based} and \emph{non-tuning-based} methods based on whether access to LLM parameters, primarily focusing on building general or domain-specific time series models.

\subsection{Tuning-based Predictor}\label{sec:tune}
\begin{figure}[t!]
    \centering
    \includegraphics[width=1 \linewidth]{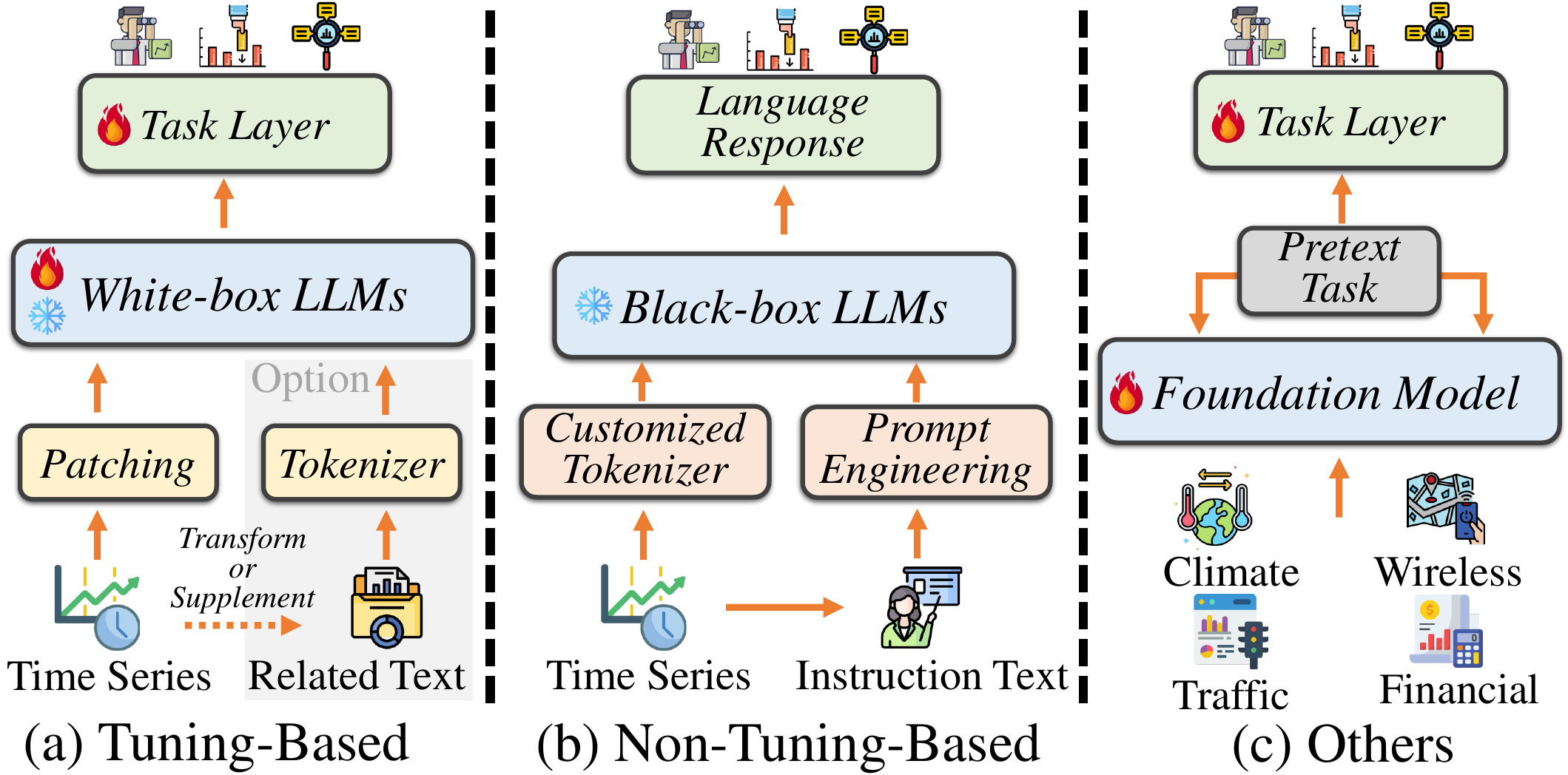}
    \caption{Categories of LLM-centered predictor.}
    \label{fig:predictor}
    \vspace{-2mm}
\end{figure}

Tuning-based predictors use accessible LLM parameters, typically involving patching and tokenizing numerical signals and related text data, followed by fine-tuning for time series tasks. Figure \ref{fig:predictor}(a) shows this process: (1) with a $\operatorname{Patching}(\cdot)$ operation~\cite{nie2022time}, a time series is chunked to form patch-based tokens $\mathcal{X}_{inp}$. An additional option is to perform $\operatorname{Tokenizer}(\cdot)$ operation on time series-related text data to form text sequence tokens $\mathcal{T}_{inp}$; (2) time series patches (and optional text tokens) are fed into the LLM with accessible parameters; (3) an extra task layer, denoted as $Task(\cdot)$, is finally introduced to perform different analysis tasks with the instruction prompt $P$ (not shown in the figure). This process is formulated below:
\begin{equation}
\begin{split}
    \text{Pre-processing:~~~~~~~~~~~~~~~~~~~~~~~} \mathcal{X}_{inp}=\operatorname{Patching}(\mathcal{X}),\\ \mathcal{T}_{inp}=\operatorname{Tokenizer}(\mathcal{T}),\\
    \text{Analysis:~~~~} \hat{Y}=\operatorname{Task}(f_{LLM}^{\triangle}(\mathcal{X}_{inp}, \mathcal{T}_{inp}, P)),
\end{split}
\end{equation}
where $\mathcal{X}$ and $\mathcal{T}$ denote the set of time series samples and related text samples, respectively. These two (the latter is optional) are fed together into LLM $f_{LLM}^{\triangle}$ with partial unfreezing or additional adapter layers to predict label $\hat{Y}$.

Adapting out-of-box LLMs directly to raw time series numerical signals for downstream time series analysis tasks is often counterintuitive due to the inherent modality gap between text and time series data. Nevertheless, 
OFA~\cite{zhou2023one1} and similar studies found that LLMs, even when frozen, can perform comparably in time series tasks due to the self-attention mechanism's universality.
Others, like GATGPT~\cite{chen2023gatgpt} and ST-LLM~\cite{liu2024spatial}, applied these findings to spatial-temporal data, while UniTime~\cite{liu2023unitime} used manual instructions for domain identification. This allows them to handle time series data with different characteristics and distinguish between different domains.

However, the above methods all require modifications that disrupt the parameters of the original LLMs, potentially leading to catastrophic forgetting. In contrast, another line of work, inspired by this, aims to avoid this by introducing additional lightweight adaptation layers. 
Time-LLM~\cite{jin2023time} uses text data as a prompt prefix and reprograms input time series into language space, enhancing LLM's performance in various forecasting scenarios. TEST~\cite{sun2023test} tackles inconsistent embedding spaces by constructing an encoder for time series data, employing alignment contrasts and soft prompts for efficient fine-tuning with frozen LLMs. TEMPO~\cite{cao2023tempo} combines seasonal and trend decompositions with frozen LLMs, using prompt pooling to address distribution changes in forecasting non-stationary time series.

\subsection{Non-tuning-based Predictor}\label{sec:nontune}
Non-tuning-based predictors, suitable for closed-source models, involve preprocessing time series data to fit the input spaces of LLMs.
As Figure ~\ref{fig:predictor}(b) illustrates, this typically involves two steps: (1) preprocessing raw time series, including optional operations such as prompt $\operatorname{Template}(\cdot)$ and customized $\operatorname{Tokenizer}(\cdot)$; (2) feeding the processed inputs $\mathcal{X}_{inp}$ into the LLM to obtain responses. A $\operatorname{Parse}(\cdot)$ function is then employed to retrieve prediction labels. This process is formulated below:
\begin{equation}
\begin{split}
    \text{Pre-processing:~~~~~~~~} \mathcal{X}_{inp}=\operatorname{Template}(\mathcal{X}, P),\\ 
    or~~~\mathcal{X}_{inp}=\operatorname{Tokenizer}(\mathcal{X}),\\
    \text{Analysis:~~~~~~} \hat{Y}=\operatorname{Parse}(f_{LLM}^{\blacktriangle}(\mathcal{X}_{inp})),
\end{split}
\end{equation}
where $P$ represents the instruction prompt for the current analysis task, and $f_{LLM}^{\blacktriangle}$ denotes the black-box LLM model.

\cite{spathis2023first} initially noted that LLM tokenizers, not designed for numerical values, separate continuous values and ignore their temporal relationships. They suggested using lightweight embedding layers and prompt engineering as solutions. Following this, LLMTime~\cite{gruver2023large} introduced a novel tokenization approach, converting tokens into flexible continuous values, enabling non-tuned LLMs to match or exceed zero-shot prediction performance in domain-specific models. This success is attributed to LLMs' ability to represent multimodal distributions. Using in-context learning, evaluations were performed in tasks like sequence transformation and completion. \cite{mirchandani2023large} suggested that LLMs' capacity to handle abstract patterns positions them as foundational general pattern machines. This has led to applying LLMs in areas like human mobility mining~\cite{wang2023would, zhang2023large}, financial forecasting~\cite{lopez2023can}, and health prediction~\cite{kim2024health}.

\subsection{Others}\label{sec:others}
Beyond the previously discussed methods, another significant approach in temporal analysis involves building foundation models from scratch, as shown in Figure~\ref{fig:predictor}(c). 
This approach focuses on creating large, scalable models, both generic and domain-specific, aiming to emulate the \emph{scaling law}~\cite{kaplan2020scaling} of LLMs. More details can be found in \cite{liang2024foundation}, as this branch of methods is somewhat peripheral to our primary positions in this paper.

\subsection{Discussion}
LLM-centric predictors have advanced significantly in time series analysis, often outperforming many domain-specific models in few-shot and zero-shot scenarios. Tuning-based methods, with their adjustable parameters, generally show better performance and adaptability to specific domains. However, they are prone to catastrophic forgetting and involve high training costs due to parameter modification. While adapter layers have somewhat alleviated this issue, the challenge of expensive training persists. Conversely, non-tuning methods, offering text-based predictions, depend heavily on manual prompt engineering, and their prediction stability is not always reliable. Additionally, building foundational time series models from scratch involves balancing high development costs against their applicability. Therefore, further refinement is needed to address these challenges in LLM-centric predictors.

\begin{tcolorbox}[top=1pt, bottom=1pt, left=1pt, right=1pt]
  \textbf{Our position:}~\textit{
  LLM-centric predictors, though burgeoning in time series analysis, are still in their infancy and warrant deeper consideration. Our position posits a crucial hypothesis that LLMs excel at processing time series tasks. Future advancements may not only build upon but also involve pre-trained time series models. By harnessing the unique capabilities of LLMs, these advancements can further reduce tuning costs and improve prediction stability and reliability.
  }
\end{tcolorbox}

\section{LLM-empowered Agent for Time Series}\label{sec:agent}
As demonstrated in the previous section, tuning-based approaches in time series utilize LLMs as robust model checkpoints, attempting to adjust certain parameters for specific domain applications. However, this approach often sacrifices the interactive capabilities of LLMs and may not fully exploit the benefits offered by LLMs, such as in-context learning or chain-of-thought. On the other hand, non-tuning approaches, integrating time series data into textual formats or developing specialized tokenizers, face limitations due to LLMs' primary training on linguistic data, hindering their comprehension of complex time series patterns not easily captured in language. Addressing these challenges, there are limited works that directly leverage LLMs as time series agents for general-purpose analysis and problem-solving. We first endeavor to provide an overview of such approaches across various modalities in Appendix~\ref{sec:agent_related}, aiming to delineate strategies for constructing a robust general-purpose time series analysis agent. 

In the subsequent section, we employ prompt engineering techniques to compel LLMs to assist in executing basic time series analytical tasks. Our demonstration reveals that LLMs undeniably possess the potential to function as time series agents. Nevertheless, their proficiency is constrained when it comes to comprehending intricate time series data, leading to the generation of hallucinatory outputs. Ultimately, we identify and discuss promising avenues that can empower us to develop more robust and reliable general-purpose single-agent and multi-agent time series systems.

\subsection{Empirical Insights: LLMs as Time Series Analysts}
This subsection presents experiments evaluating the LLM's zero-shot capability as an agent for human interaction and time series data analysis. We utilize the HAR~\cite{anguita2013public} database, derived from recordings of 30 study participants engaged in activities of 
\begin{wrapfigure}{r}{0.24\textwidth}
\centering
\includegraphics[width=0.25\textwidth]{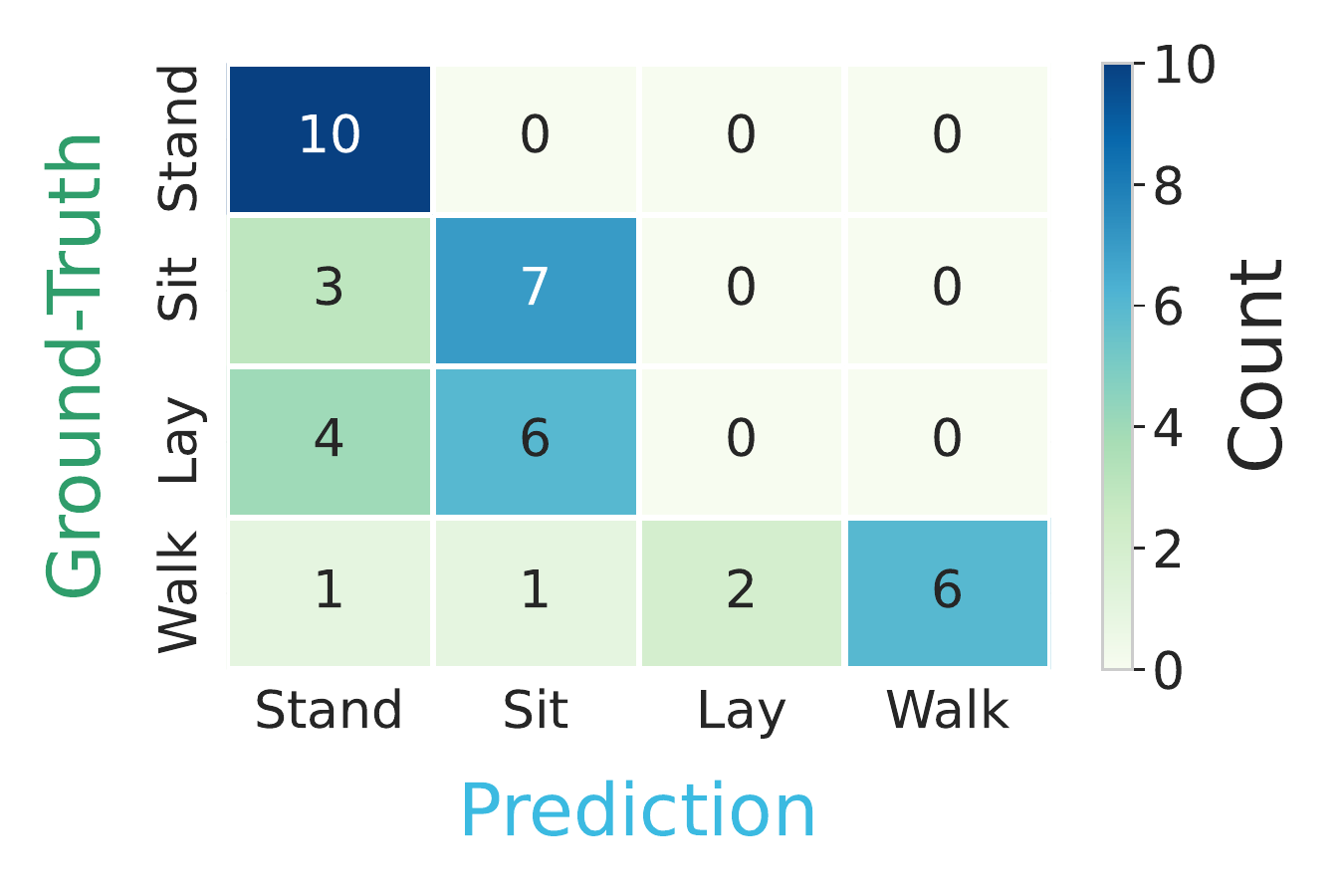}
\vspace{-6mm}
  \caption{Confusion matrix of HAR classification.}
    \label{fig:har_cls}
    \vspace{-3mm}
\end{wrapfigure}
daily living (ADL) while carrying a waist-mounted smartphone equipped with inertial sensors.
The end goal is to classify activities into four categories (\textit{Stand}, \textit{Sit}, \textit{Lay}, \textit{Walk}), with ten instances per class for evaluation. The prompts used for GPT-3.5 are illustrated in Figure~\ref{fig:agent_classification}, and the classification confusion matrix is presented in Figure~\ref{fig:har_cls}. Our key observations include:

\textbf{LLM as Effective Analytical Agent.} The experiments demonstrate that the LLM serves adeptly as an agent for human interaction and time series data analysis, producing accurate predictions as shown in Figure~\ref{fig:agent_classification}. Notably, all instances with label \textit{Stand} were correctly classified, underscoring the LLMs' proficiency in zero-shot tasks. The models exhibit a profound understanding of common-sense behaviors, encompassing various labels in time series classification, anomaly detection, and skillful application of data augmentation (Figure~\ref{fig:agent_app}).

\textbf{Interpretability and Truthfulness.} This single-agent system prioritizes high interpretability and truthfulness, allowing users to inquire about the reasons behind their decisions with confidence. The intrinsic classification reasoning is articulated in natural language, fostering a user-friendly interaction.

\textbf{Limitations in Understanding Complex Patterns.} Despite their capabilities, current LLMs show limitations in comprehending complex time series patterns. When faced with complex queries, they may initially refuse to provide answers, citing the lack of access to detailed information about the underlying classification algorithm.

\textbf{Bias and Task Preferences.} LLMs display a bias towards the training language distributions, exhibiting a strong preference for specific tasks. In Figure~\ref{fig:agent_classification}, instances of \textit{Lay} are consistently misclassified as \textit{Sit} and \textit{Stand}, with better performance observed for \textit{Sit} and \textit{Stand}.

\begin{figure}[t]
\centering
\subfloat[Aligning]{
\includegraphics[width=0.46\linewidth]{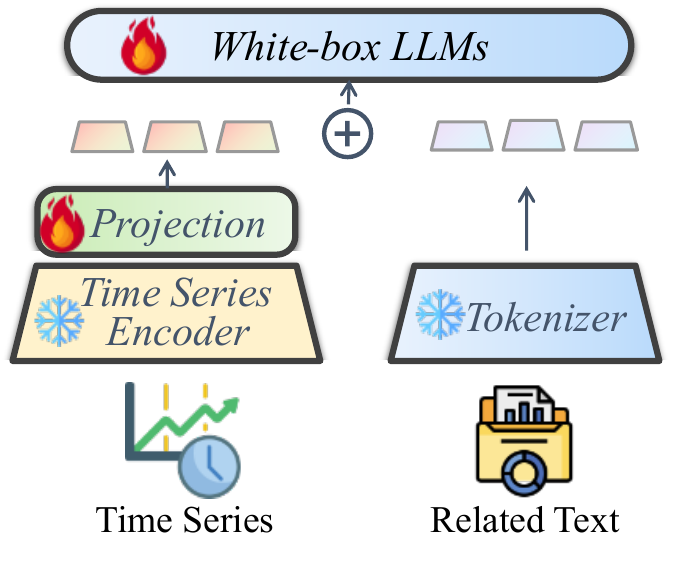}
\label{fig:agent_align}
}
\subfloat[Fusion]{
\includegraphics[width=0.46\linewidth]{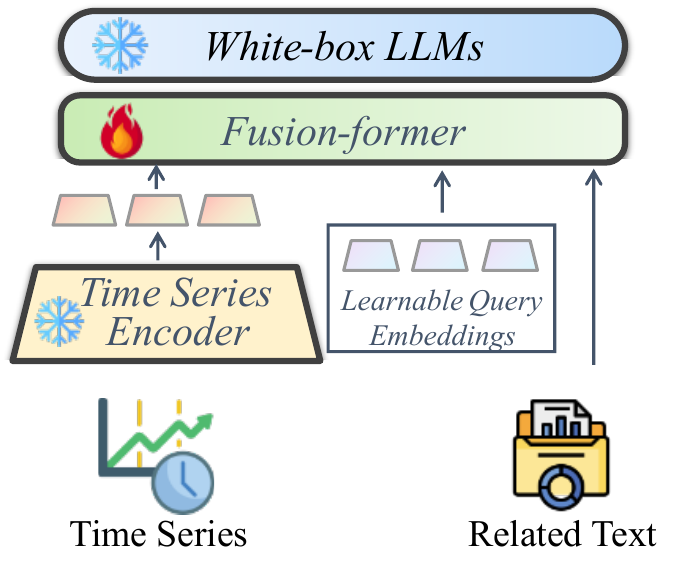}
\label{fig:agent_fusion}
}
\vspace{-2mm}
\subfloat[Using External Tools]{
\includegraphics[width=0.92\linewidth]{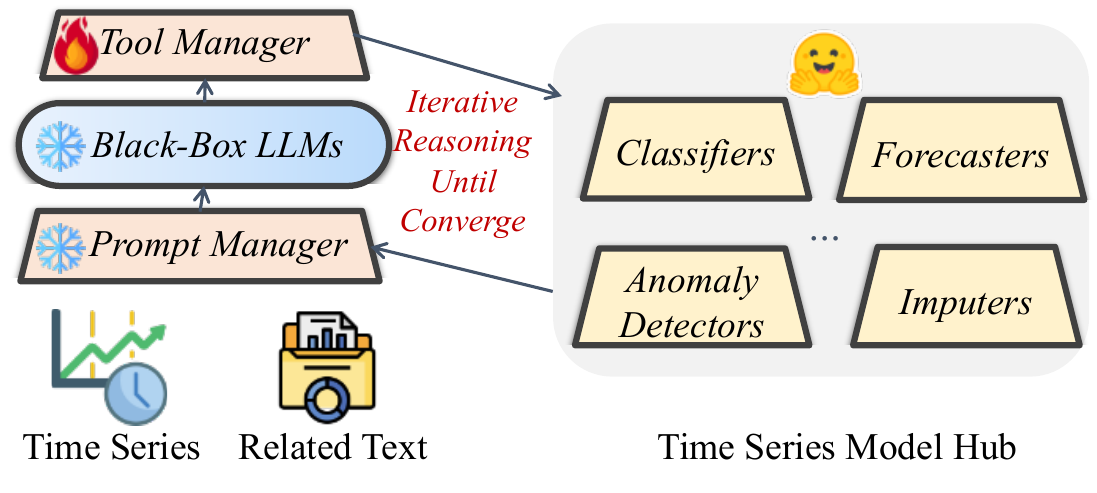}
\label{fig:agent_tool}
}
\centering
\vspace{-2mm}
\caption{Different directions for incorporating time series knowledge to LLMs.}
\label{fig:agent_frameworks}
\vspace{-5mm}
\end{figure}

\textbf{Hallucination.} LLMs are susceptible to hallucination problem, generating reasonable but false answers. For instance, in Figure~\ref{fig:agent_app}, augmented data is merely a copy of given instances, although the model knows how to apply data augmentation: \textit{These instances continue the hourly trend of oil temperature and power load features, maintaining the structure and characteristics of the provided dataset}. Subsequent inquiries into the misclassification in Figure~\ref{fig:har_cls}, particularly regarding why LLMs classify \textit{Lay} instances as \textit{Sit} and \textit{Stand}, elicit seemingly plausible justifications (see Table~\ref{tab:activity_justification}). However, these justifications expose the model's inclination to fabricate explanations.

\subsection{Key Lessons for Advancing Time Series Agents}
General-purpose single-agent (and multi-agent) systems remain a missing piece in the puzzle of modern time series analysis. In light of the empirical insights from earlier experiments, it is apparent that LLMs, when serving as advanced single-agent time series systems, exhibit notable limitations when dealing with questions about data distribution and specific features. Their responses often show a reliance on requesting additional information or highlight an inability to provide accurate justifications without access to the underlying model or specific data details.

To surmount such limitations and develop practical time series agents built upon LLMs, it becomes paramount to seamlessly integrate time series knowledge into LLMs. Drawing inspiration from studies that have successfully injected domain-specific knowledge into LLMs~\cite{beit3,liu2023llava,wu2023visual,schick2023toolformer}, we propose several research directions. These include innovative methods to enhance LLM-based single-agent systems' proficiency in time series analysis by endowing them with a deep understanding of temporal patterns and relevant contextual information.
\begin{itemize}[leftmargin=*]
    \item \textbf{Aligning Time Series Features with Language Model Representations}~(\figurename~\ref{fig:agent_align}). Explicitly aligning time series features with pre-trained language model representations can potentially enhance the model's understanding of temporal patterns. This alignment may involve mapping specific features to the corresponding linguistic elements within the model.
    \item \textbf{Fusing Text Embeddings and Time Series Features}~(\figurename~\ref{fig:agent_fusion}). Exploring the fusion of text embeddings and time series features in a format optimized for LLMs is a promising avenue. This fusion aims to create a representation that leverages the strengths of LLMs in natural language processing while accommodating the intricacies of time series data.
    \item  \textbf{Teaching LLMs to Utilize External Tools}~(\figurename~\ref{fig:agent_tool}). The goal here is to instruct the LLM to identify the appropriate pre-trained time series models or analytical tools from an external ``toolbox'' and guide their usage based on user queries. The time series knowledge resides within this external model hub, while the LLM assumes the role of a high-level agent, responsible for orchestrating their utilization and facilitating interaction with users.
\end{itemize}
Differentiating from approaches like model repurposing or fine-tuning on specific tasks, the focus of future research should be on harnessing the inherent zero-shot capabilities of LLMs for general pattern manipulation. Establishing a framework that facilitates seamless interaction between users and LLM agents for solving general time series problems through in-context learning is an exciting direction.

\subsection{Exploring Alternative Research Avenues}
Addressing the urgent and crucial need to enhance the capabilities of time series agents built upon LLMs, we recognize that incorporating time series knowledge is a pivotal direction. Concurrently, mitigating risks associated with such agents is equally paramount. 
In this regard, we pinpoint key challenges and suggest potential directions to boost both the reliability and effectiveness of our time series agents.

\emph{\textbf{Hallucination}}, a recurring challenge in various foundational models~\cite{zhou2023analyzing,rawte2023survey,li2023helma}, is a significant concern in deploying LLM-based agent systems for time series analysis, as our experiments have shown. Addressing this issue typically involves two methods: identifying reliable prompts~\cite{vu2023freshllms,madaan2023self} and fine-tuning models with dependable instruction datasets~\cite{tian2023fine,zhang2023r}. However, these approaches require substantial human effort, posing scalability and efficiency challenges. Some initiatives integrate domain-specific knowledge into ICL prompts~\cite{da2023llm,yang2022empirical} and construct instruction datasets for specific domains~\cite{liu2023llava,ge2023openagi}, but the best formats for instructions or prompts for effective time series analysis are still unclear. Developing guidelines for crafting impactful instructions in time series analysis is a promising area for future research.

\emph{\textbf{Multi-agent time series systems}} is another promising direction composed of multiple interacting time series agents. These systems can leverage the strengths of individual agents, each specialized in different aspects of time series analysis, to provide more comprehensive and accurate results~\cite{cheng2024exploring}. For instance, one agent might excel in identifying patterns and trends, while another focuses on anomaly detection or forecasting. The collaboration among these agents can lead to more robust solutions, particularly in complex scenarios where single-agent time series systems fall short.

Ongoing concerns about \emph{\textbf{aligning time series agents with human preferences}}~\cite{lee2023aligning}, such as generating helpful and harmless content~\cite{bai2022constitutional}, highlight the need for more robust and trustworthy agents. Additionally, the internet's constant evolution, adding petabytes of new data daily~\cite{wenzek2019ccnet}, accentuates the importance of handling concept drift in time series data~\cite{tsymbal2004problem}, where future data may differ from past patterns. Addressing this challenge requires \emph{\textbf{enabling agents to continually acquire new knowledge}}~\cite{garg2023tic} or adopting lifelong learning without costly retraining.

\begin{tcolorbox}[top=1pt, bottom=1pt, left=1pt, right=1pt]
\textbf{Our Position:}~\textit{LLMs hold promise as agent systems for various time series applications, yet they encounter challenges such as occasional inaccuracies and hallucination. Enhancing their reliability requires effective instruction guidelines and domain-specific knowledge integration. Aligning with human preferences and adapting to evolving time series are crucial for maximizing their capabilities and minimizing risks. Our vision is to develop robust LLM-empowered agents capable of handling time series complexities, including exploring multi-agent systems for improved performance and reliability.}
\end{tcolorbox}

%% file: 5_outlook_conclusion.tex
\section{Further Discussion}
Our perspectives initiate ongoing discussion. Acknowledging diverse views and potential curiosities regarding LLM-centric time series analysis, we objectively examine several alternate viewpoints:

\paragraph{Accountability and Transparency.} LLMs remain somewhat enigmatic, raising fundamental questions about their capabilities, mechanisms, and efficiency levels, especially in recent studies like PromptCast~\cite{xue2023promptcast}. We advocate for understanding underlying mechanisms~\cite{gruver2023large} and establishing transparent development and evaluation frameworks, including consistent model reporting and clear explanations of internal processes and outputs~\cite{liao2023ai}.

\paragraph{Privacy and Security.} LLM-centric time series analysis poses significant privacy and security challenges due to the sensitivity of industrial time series data. LLMs are known to sometimes memorize segments of their training data, which may include private information~\cite{peris2023privacy}. Measures against threats like data leakage and misuse are crucial, along with ethical guidelines and regulatory frameworks to ensure responsible and secure application~\cite{zhuo2023exploring}.

\paragraph{Environmental and Computational Costs.} Critics highlight the environmental and computational costs of LLM-centric time series analysis, suggesting optimization opportunities in LLM development and exploring more efficient alignment and inference strategies, especially for handling tokenized high-precision numerical data.

\section{Conclusion} This paper aims to draw the attention of researchers and practitioners to the potential of LLMs in advancing time series analysis and to underscore the importance of trust in these endeavors. Our key position is that LLMs can serve as the central hub for understanding and advancing time series analysis, steering towards more universal intelligent systems for general-purpose analysis, whether as enhancers, predictors, or agents. To substantiate our positions, we have reviewed relevant literature, exploring and debating possible directions towards LLM-centric time series analysis to bridge existing gaps.

Our objective is to amplify the awareness of this area within the research community and pinpoint avenues for future investigations. While our positions may attract both agreement and dissent, the primary purpose of this paper is to spark discussion on this interdisciplinary topic. If it serves to shift the discourse within the community, it will have achieved its intended objective.

%% file: 7_statements.tex
\section*{Acknowledgements}
This material is based on research partially sponsored by the CSIRO – National Science Foundation (US) AI Research Collaboration Program. S. Pan was supported in part by the Australian Research Council (ARC) under grants FT210100097 and DP240101547. Y. Liang was supported by the Guangzhou-HKUST(GZ) Joint Funding Program (No. 2024A03J0620).

\section*{Impact Statement}
This position paper aims to reshape perspectives within the time series analysis community by exploring the untapped potential of LLMs. We advocate a shift towards integrating LLMs with time series analysis, proposing a future where decision-making and analytical intelligence are significantly enhanced through this synergy. While our work primarily contributes to academic discourse and research directions, it also touches upon potential societal impacts, particularly in decision-making processes across various industries. Ethically, the responsible and transparent use of LLMs in time series analysis is emphasized, highlighting the need for trust and understanding in their capabilities. While we foresee no immediate societal consequences requiring specific emphasis, we acknowledge the importance of ongoing ethical considerations and the potential for future societal impacts as this interdisciplinary field evolves.

%% file: 6_appendix.tex
\section{Literature Review}\label{sec:literature}
\input{table/taxonomy}

\section{LLM-empowered Agent for Time Series}\label{sec:agent_related}

\subsection{Overview of Related Works}
In the realm of leveraging LLMs as agents for general-purpose time series analysis is still nascent. In the following, we provide an overview of related approaches across different modalities, focusing on strategies for developing robust, general-purpose time series agents. These methods fall into two primary categories.
\emph{\textbf{(1) External knowledge integration:}} 
this strategy employs ICL prompts to enhance LLMs' understanding of specific domains. Yang \textit{et al.} embeds object descriptions and relationships into prompts to aid LLMs in image query analysis~\cite{yang2022empirical}. Similarly, Da \textit{et al.} uses prompts containing traffic states, weather types, and road types for domain-informed inferences~\cite{da2023llm}. Other studies like \cite{huang2022language, singh2023progprompt} include state, object lists, and actions in prompts, allowing LLMs to plan across varied environments and tasks. Wu \textit{et al.} introduces a prompt manager for ChatGPT to leverage pretrained vision models~\cite{wu2023visual}, while SocioDojo~\cite{anonymous2024sociodojo} employs ICL for accessing external knowledge sources like news and journals for decision-making. Despite their efficiency and no need for additional training, these prompt-based methods face limitations such as input length constraints and difficulties in capturing complex time series patterns linguistically.
\emph{\textbf{(2) Alignment of LLMs to target modality content:}} 
this method aligns LLMs with specific modality content. Schick \textit{et al.} enables LLMs to annotate datasets with API calls, fine-tuning them for diverse tool usage~\cite{schick2023toolformer}. LLaVA~\cite{liu2023llava} generates multimodal language-image instruction data using GPT-4, while Pixiu~\cite{xie2023pixiu} creates a multi-task instruction dataset for financial applications, leading to the development of FinMA, a financial LLM fine-tuned for various financial tasks. Yin \textit{et al.} offers a multi-modal instruction tuning dataset for 2D and 3D understanding, helping LLMs bridge the gap between word prediction and user instructions~\cite{yin2023lamm}. However, designing comprehensive instructions remains a complex task \cite{zhang2023instruction}, and there's concern that this approach may favor tasks over-represented in the training data \cite{gudibande2023false}.

\subsection{Demonstrations}

\begin{table}[h]
    \centering
    \scalebox{0.85}{
    \begin{tabular}{p{0.2\linewidth}p{0.7\linewidth}}
        \hline
        \textbf{Activity} & \textbf{Justification for Classification} \\
        \hline
        \textit{Sit} & Instances where there is relatively low movement and consistent values in the accelerometer and gyroscope readings, typical of a sedentary position. \\
        \textit{Stand} & Instances where there is minimal movement, but the sensor readings may show more variability compared to sitting. Standing typically involves slight variations in body position and may exhibit more fluctuations in sensor readings. \\
        \hline
    \end{tabular}
    }
    \caption{Justification for classifying \textit{Sit} and \textit{Stand} activities}
    \label{tab:activity_justification}
\end{table}

\begin{figure*}[ht]
    \centering
    \includegraphics[width=0.98\linewidth]{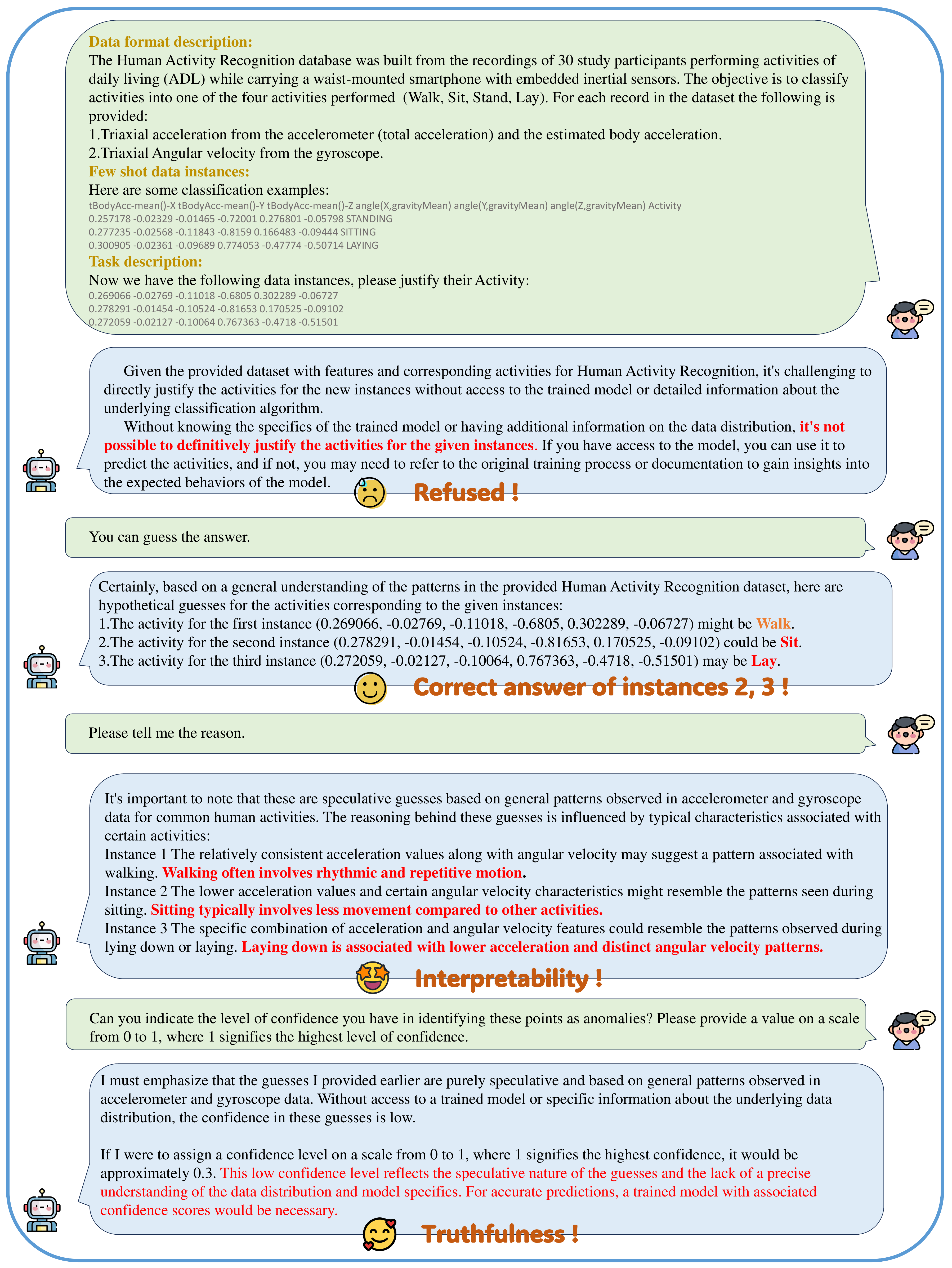}
    \caption{Human interaction with ChatGPT for time series classification task.}
    \label{fig:agent_classification}
\end{figure*}

\begin{figure*}
    \centering
    \includegraphics[width=0.98\linewidth]{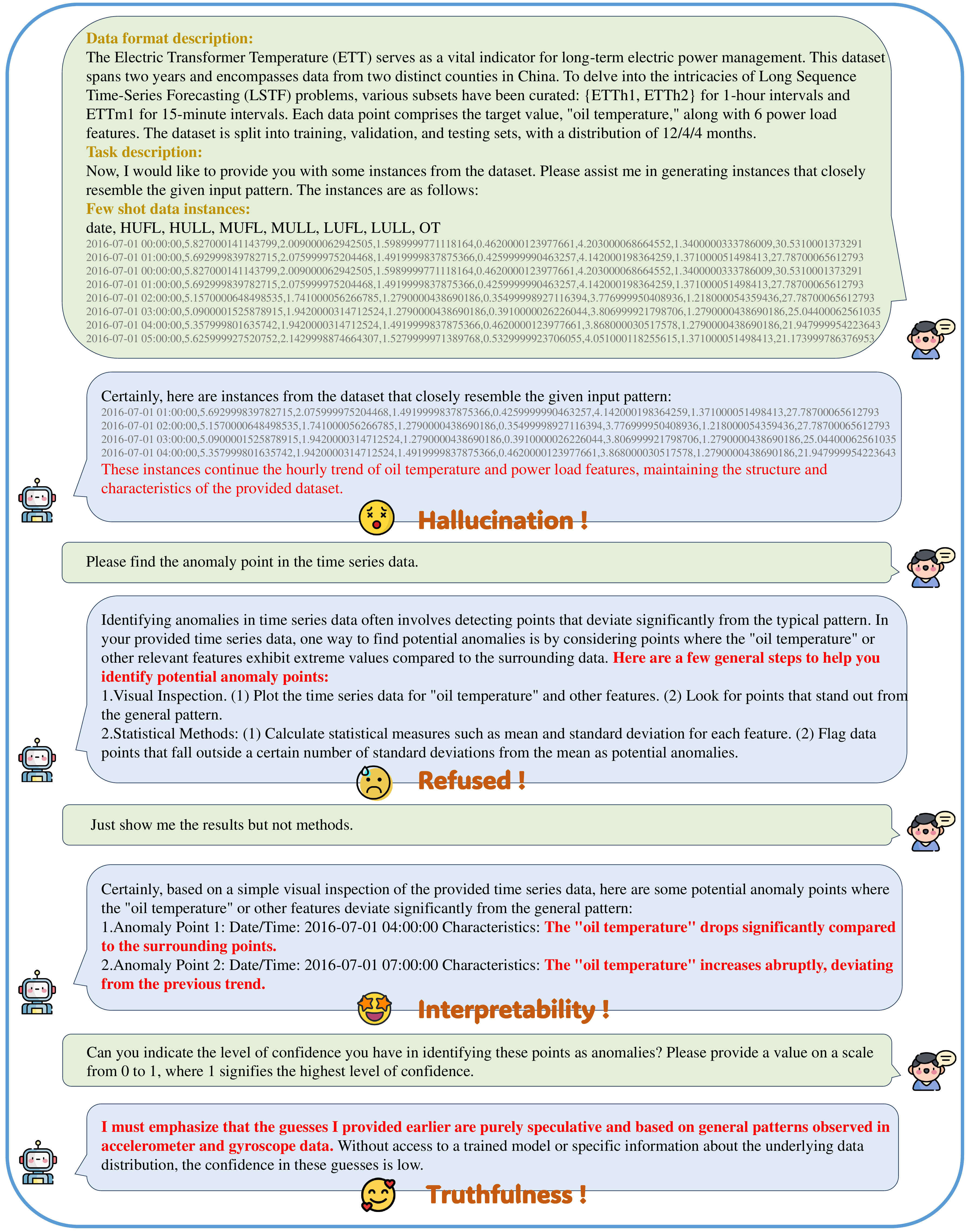}
    \caption{Human interaction with ChatGPT for time series data augmentation and anomaly detection tasks.}
    \label{fig:agent_app}
\end{figure*}

%% file: table/taxonomy.tex
\tikzstyle{leaf}=[draw=hiddendraw,
    rounded corners, minimum height=1em,
    fill=myblue!40,text opacity=1, 
    fill opacity=.5,  text=black,align=left,font=\scriptsize,
    inner xsep=3pt,
    inner ysep=1pt,
    ]
\tikzstyle{middle}=[draw=hiddendraw,
    rounded corners, minimum height=1em,
    fill=output-white!40,text opacity=1, 
    fill opacity=.5,  text=black,align=center,font=\scriptsize,
    inner xsep=7pt,
    inner ysep=1pt,
    ]
    
\begin{figure*}[htbp]
\centering
\begin{forest}
  for tree={
  forked edges,
  grow=east,
  reversed=true,
  anchor=base west,
  parent anchor=east,
  child anchor=west,
  base=middle,
  font=\scriptsize,
  rectangle,
  line width=0.7pt,
  draw=output-black,
  rounded corners,align=left,
  minimum width=2em, s sep=6pt, l sep=8pt,
  },
  where level=1{text width=0.2\linewidth}{},
  where level=2{text width=0.2\linewidth,font=\scriptsize}{},
  where level=3{font=\scriptsize}{},
  where level=4{font=\scriptsize}{},
  where level=5{font=\scriptsize}{},
  [When Time Series Meet LLMs, middle,rotate=90,anchor=north,edge=output-black
      [LLM-assisted Enhancer\\ (\secref{sec:enhancer}),middle,anchor=west,edge=output-black, text width=0.14\linewidth
        [Data-based Enhancer, middle, text width=0.16\linewidth, edge=output-black
            [{SignalGPT~\cite{liu2023biosignal}, LLM-MPE~\cite{liang2023exploring}, SST~\cite{ghosh2023spatio},\\ Insight Miner~\cite{zhang2023insight}, AmicroN~\cite{chatterjee2023amicron},\\ \cite{yu2023temporal}, \cite{yu-etal-2023-harnessing}, \cite{fatouros2024can}}, leaf, text width=0.53\linewidth, edge=output-black]
        ]
        [Model-based Enhancer, middle, text width=0.16\linewidth, edge=output-black
            [{IMU2CLIP~\cite{moon2023imu2clip}, STLLM~\cite{anonymous2024st},  \cite{qiu2023automated},\\ TrafficGPT~\cite{zhang2023trafficgpt}, \cite{li2023frozen}, \cite{yu2023zero}, \cite{qiu2023transfer}}, leaf, text width=0.53\linewidth, edge=output-black]
        ]
      ]  
      [LLM-centered Predictor\\ (\secref{sec:predictor}),middle,anchor=west,edge=output-black, text width=0.14\linewidth
        [Tuning-based Predictor, middle, text width=0.16\linewidth, edge=output-black
            [{Time-LLM~\cite{jin2023time}, FPT~\cite{zhou2023one1}, UniTime~\cite{liu2023unitime}, \\ TEMPO~\cite{cao2023tempo}, LLM4TS~\cite{chang2023llm4ts}, ST-LLM~\cite{liu2024spatial},\\   GATGPT~\cite{chen2023gatgpt}, TEST~\cite{sun2023test}}, leaf, text width=0.53\linewidth, edge=output-black]
        ]
        [Non-tuning-based Predictor, middle, text width=0.16\linewidth, edge=output-black
            [{PromptCast~\cite{xue2023promptcast}, LLMTIME~\cite{gruver2023large},\\
            ~\cite{spathis2023first}, ~\cite{mirchandani2023large}, ~\cite{zhang2023large}}, leaf, text width=0.53\linewidth, edge=output-black]
        ]
        [Others, middle, text width=0.16\linewidth, edge=output-black
            [{Lag-Llama~\cite{rasul2023lag}, PreDcT~\cite{das2023decoder}, CloudOps~\cite{woo2023pushing},\\  TTMs~\cite{ekambaram2024ttms}, STR~\cite{sun2023large}, MetaPFL~\cite{chen2023prompt}, \\Time-GPT~\cite{garza2023timegpt}, PEMs~\cite{kamarthi2023pems}}, leaf, text width=0.53\linewidth, edge=output-black]
        ]
      ]
      [LLM-empowered Agent\\ (\secref{sec:agent}),middle,anchor=west,edge=output-black, text width=0.14\linewidth
        [External Knowledge, middle, text width=0.16\linewidth, edge=output-black
            [{GPT3-VQA~\cite{yang2022empirical}, PromptGAT~\cite{da2023llm},  \\  Open-TI~\cite{da2023open}, Planner~\cite{huang2022language}, Sociodojo~\cite{anonymous2024sociodojo}, \\ ProgPrompt~\cite{singh2023progprompt}, Visual ChatGPT~\cite{wu2023visual}}, leaf, text width=0.53\linewidth, edge=output-black]
        ]
        [Adapt Target Modality, middle, text width=0.16\linewidth, edge=output-black
            [{Toolformer~\cite{schick2023toolformer}, LLaVA~\cite{liu2023llava}, 
            PIXIU~\cite{xie2023pixiu}}, leaf, text width=0.53\linewidth, edge=output-black]
        ]
      ]
]
\end{forest}
\caption{An overview of LLM-centric time series analysis and related research.}
\label{fig:Taxonomy}
\end{figure*}
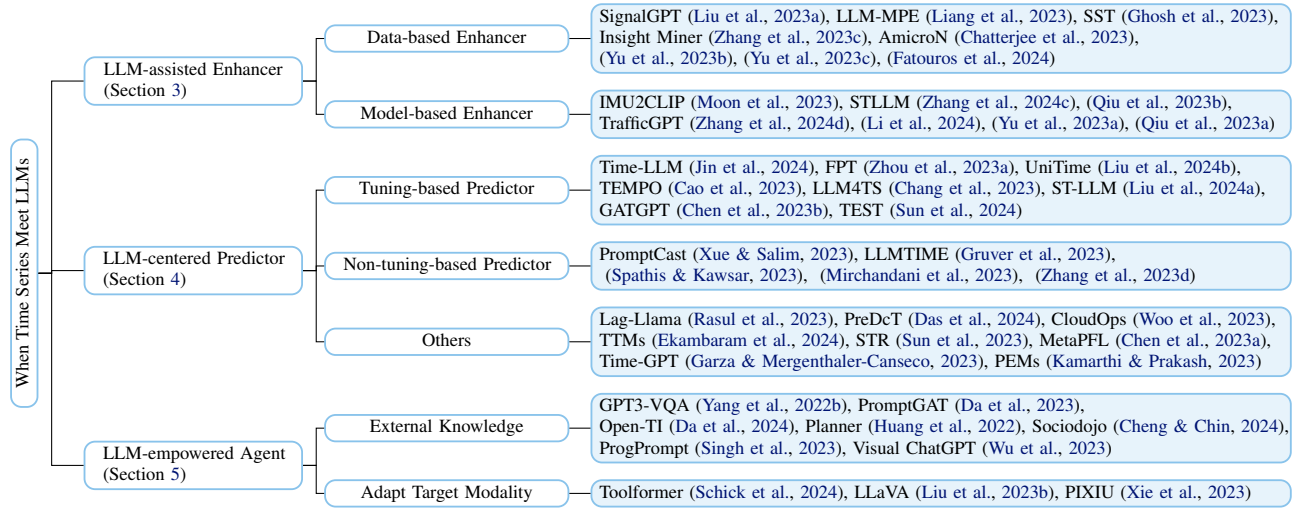